# Perceptually Optimized Image Rendering

Valero Laparra[1,2,*], Alex Berardino[2], Johannes Ballé[2], and Eero P. Simoncelli[2,3]

[1]*Image Processing Laboratory, Universitat de València, 46980 Paterna, Spain*
[2]*Center for Neural Science, New York University, New York, NY 10003, USA*
[3]*Courant Institute of Mathematical Sciences, New York University, New York, NY 10003, USA*
[*]*Corresponding author: valero.laparra@uv.es*

January 23, 2017

## Abstract

We develop a framework for rendering photographic images, taking into account display limitations, so as to optimize perceptual similarity between the rendered image and the original scene. We formulate this as a constrained optimization problem, in which we minimize a measure of perceptual dissimilarity, the Normalized Laplacian Pyramid Distance (NLPD), which mimics the early stage transformations of the human visual system. When rendering images acquired with higher dynamic range than that of the display, we find that the optimized solution boosts the contrast of low-contrast features without introducing significant artifacts, yielding results of comparable visual quality to current state-of-the art methods with no manual intervention or parameter settings. We also examine a variety of other display constraints, including limitations on minimum luminance (black point), mean luminance (as a proxy for energy consumption), and quantized luminance levels (halftoning). Finally, we show that the method may be used to enhance details and contrast of images degraded by optical scattering (e.g., fog).

## 1 Introduction

A general goal in designing a pipeline for the capture and display of photographic images is to remain as faithful to the original source as possible, minimizing distortions introduced by the sensor, coding, transmission, or display processes. If images are meant for presentation to human observers, distortion should be measured accordingly, penalizing errors that are most visually noticeable and/or disturbing, while permitting those that are perceptually unnoticeable. This strategy is most evident in the handling of color, in which both sensors and displays are designed so as to capture and render the three-dimensional subspace of wavelengths relevant for human trichromatic visual representation, while allowing significant distortion outside of this subspace.

Arguably the most significant limitations of current sensors and displays are with regard to dynamic range. Early sensors were restricted to capturing a limited luminance range, and were unable to adequately capture the majority of realistic natural scenes, which contain luminances spanning many orders of magnitude. In contrast, the human visual system is capable of sensing fixed scenes with a range of over 5 orders of magnitude in real time, and up to 8 orders of magnitude when the effects of extended temporal adaptation mechanisms are incorporated [12] (see Fig. 1). The dynamic range of sensors has steadily improved, and current sensors (often augmented with software solutions that fuse images captured at different exposures) are capable of acquiring images with luminance ranges approximating those of human vision. Despite this, even our best display devices are limited to a significantly lower dynamic range than these sensors can capture.

The simplest solution to the problem of displaying high dynamic range (HDR) images on a low dynamic range (LDR) rendering device is to linearly rescale the luminance values recorded by the sensor into the display's reproducible range of luminances. This, however, produces images that look nothing like the original scene - typically all of the low-luminance information is lost. A variety of "tone-mapping" methods have been proposed to solve this problem by nonlinearly remapping the intensities of the original image into the output range, in a way that remains faithful to the visual appearance of the original scene [29]. Most of these are based on heuristics, and require manual adjustment of parameters for best results. In addition, many display applications introduce constraints other than global luminance range, such as discrete luminance levels (i.e. halftoning), average energy consumption, as well as interactions between pixel values over space or time. Separate methods have been developed for solving each of these problems.

Here, we formulate the general problem of perceptually-accurate display rendering as a constrained optimization problem, optimizing the rendered image to minimize perceptual differences from the original, subject to any constraints imposed by the display (Fig. 1). The formulation relies on four ingredients: knowledge of the conditions of image acquisition, knowledge of the display constraints, a definition of perceptual similarity between images, and a method for optimizing the image to be rendered. In the next section, we combine these ingredients in a simple optimization problem. Because we choose a model of perceptual similarity that is continuous and differentiable, it can be efficiently solved by first-order constrained optimization techniques. We show that the solution is well defined and general, and therefore represents a framework for solving a wide class of rendering problems. Here, we restrict ourselves to grayscale images (without chromatic components) since the original perceptual model was designed for achromatic images. We show one result per experiment, more images can be found at our web page.[1]

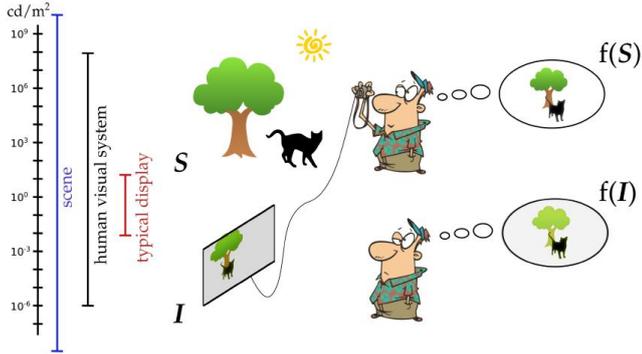

Figure 1: Perceptually optimized rendering framework. When we view a real-world scene, the luminances, specified by a vector $\boldsymbol{S}$, give rise to an internal perceptual representation $f(\boldsymbol{S})$. While luminances in the real world can range from complete darkness ($0\,cd/m^2$) to extremely bright (e.g., midday sun, roughly $10^9\,cd/m^2$), a typical display can generate a relatively narrow range of roughly 5 to $300\,cd/m^2$. The optimization goal is to adjust $\boldsymbol{I}$ so as to minimize the difference between the perceptual representations, $f(\boldsymbol{S})$ and $f(\boldsymbol{I})$, while remaining within the set of images that can be generated by the display.

In section 3, we optimize images captured under differing acquisition conditions for rendering on the same display. We start with calibrated images, where the original scene luminances are known. We also deal with the more common scenario in which the exact luminances of the original scene are unknown (the tone mapping problem). In this scenario, we have to make some educated guesses about the physical conditions of the original scene. We demonstrate the effect that different starting assumptions have on the optimized images. Moreover, we take advantage of these effects to solve other image processing problems, such as detail enhancement and haze removal, by manipulating these source assumptions. For each of these tasks, we compare the results with state-of-the-art algorithms designed to solve each specific case. In section 4, we optimize images to be displayed under differing display restrictions, including luminance limited displays, energy limited displays, and displays restricted to a small set of output values. Finally, we analyze the effect that each component of our perceptual measure has on the quality of our optimized images.

## 2 Optimal rendering framework

Optimally rendering an image, $\boldsymbol{I}$, on a given display means displaying it in such a way that it remains faithful to the human perception of the original scene, $\boldsymbol{S}$. We formalize this as a constrained optimization problem:

$$\hat{\boldsymbol{I}} = \arg\min_{\boldsymbol{I}} D(\boldsymbol{S}, \boldsymbol{I}), \quad \text{s.t. } \boldsymbol{I} \in \mathcal{C}, \qquad (1)$$

where $D(\cdot, \cdot)$ is a measure of human perceptual dissimilarity, and $\mathcal{C}$ is the set of all images that can be rendered on the display. This formulation can express many well-known rendering problems, such as tone mapping or dithering, which differ only in the specification of $\mathcal{C}$. In general, the optimization problem expressed in Eq. (1) cannot be solved analytically, and thus we will not obtain an explicit function to compute $\hat{\boldsymbol{I}}$, given $\boldsymbol{S}$ and $\mathcal{C}$. Rather than assume a functional form for this mapping, we choose a perceptual measure that is differentiable with respect to $\boldsymbol{I}$, and use modern high-dimensional optimization tools to numerically solve for the optimal $\hat{\boldsymbol{I}}$. Specifically, we descend the objective function, alternating between minimizing the perceptual distance, and projecting the image back onto the constraint set. Specific formulations for different example problems can be found online[1].

We follow a principled, two-step approach to quantify perceptual distance. Rather than defining a perceptual distance directly (as in SSIM [30], for example), we first define a nonlinear *perceptual transform* $f(\cdot)$, which approximates the computations performed within the early stages of the human visual system. We apply this to both the original scene luminances, $\boldsymbol{S}$, and the rendered image, $\boldsymbol{I}$, and then measure the distance between $f(\boldsymbol{S})$ and $f(\boldsymbol{I})$. A perceptual distance measure constructed this way is symmetric and yields a value zero for identical images, but may not satisfy the remaining mathematical requirements of a metric. Specifically, it might also yield zero for non-identical images (if the transformation discards information), and may not satisfy the triangle inequality.[2] Despite this, we refer to it casually as a metric throughout the paper.

Figure 2 illustrates the components of the perceptual transform, for which we use the Normalized Laplacian Pyramid (NLP), a multi-scale nonlinear representation that mimics the operations of the retina and lateral geniculate nucleus in the human visual system. We have previously shown that distances measured between two images in the perceptual space defined by the NLP are highly correlated with human judgments [15]. Here, we adapt this model to operate directly on luminances (rather than values that have been gamma-adjusted for a particular display), which allows use of the same units when defining constraints on acquisition and display. Luminances are first transformed elementwise using a power law, which approximate the transformation of light to voltage in retinal photoreceptors:

$$x = S^\gamma \quad (2)$$

---

[1] http://www.cns.nyu.edu/~lcv/perceptualRendering/

[2] I.e., it may technically be a *semimetric* or a *premetric*.

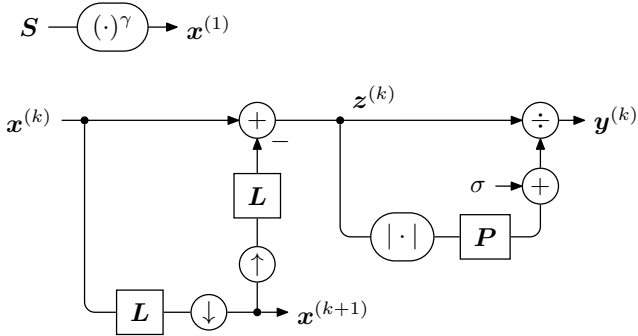
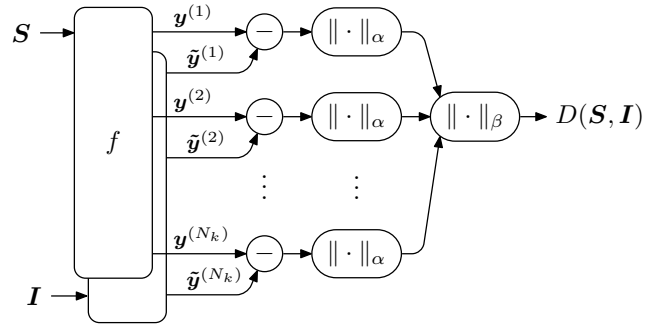

Figure 2: Perceptual transform, constructed as a Normalized Laplacian Pyramid (NLP) [15]. Scene luminances $S$ (in $cd/m^2$) are first transformed using a power function (top). The transformed luminance image is then decomposed into frequency channels, using the recursive implementation of the Laplacian Pyramid [3]. Each channel $z^{(k)}$ is then divided by a weighted sum of local amplitudes (computed with lowpass filter $P$) plus a constant, $\sigma$. Symbols ↑ and ↓ indicate upsampling and downsampling by a factor of 2, respectively.

Figure 3: Summation model, using the Normalized Laplacian Pyramid [15] as perceptual transform $f$ (see Fig. 2). For all results, we use $\alpha = 2.0$ for averaging within, and $\beta = 0.6$ for averaging across frequency channels.

This initial nonlinear transformation is followed by a recursive partition into frequency channels, as in the Laplacian Pyramid [3]:

$$\begin{aligned} x^{(k+1)} &= D\left(L(x^{(k)})\right), \\ z^{(k)} &= x^{(k)} - L\left(U(x^{(k+1)})\right), \end{aligned} \quad (3)$$

where $D(\cdot)$ and $U(\cdot)$ indicate down/up-sampling by a factor of two, respectively (figure 2). For the filtering operation $L$, we apply a spatially separable 5-tap filter, $(0.05, 0.25, 0.4, 0.25, 0.05)$, as originally specified in [3].

Within each channel, each coefficient is divided by a weighted local sum of the element-wise amplitudes (absolute values) plus a constant:

$$y^{(k)} = z^{(k)} \oslash \left(\sigma + P|z^{(k)}|\right). \quad (4)$$

where $P$ indicates convolution with a filter, and $\oslash$ indicates pointwise division. This function is a simplified variant of *divisive normalization*, used to describe the responses of neurons in different parts of the visual system [11, 26, 4]. The NLP coefficients of all channels $y^{(k)}$ combined represent the response of the perceptual transform:

$$f(S) = \{y^{(k)}; k = 1, \ldots, N_k\}. \quad (5)$$

Figure 3 illustrates the construction of the metric employed in the perceptual space. We compute the $L_\alpha$-norm of the differences between NLP coefficients within each frequency channel (that is, we raise the absolute value of each coefficient difference to the power $\alpha$, sum over the entire channel, and take the $\alpha$th root). These values are then combined across channels using an $L_\beta$-norm:

$$D(S, I) = \left[\frac{1}{N_k} \sum_{k=1}^{N_k} \left(\frac{1}{N_c^{(k)}} \sum_{i=1}^{N_c} |y_i^{(k)} - \tilde{y}_i^{(k)}|^\alpha\right)^{\frac{\beta}{\alpha}}\right]^{\frac{1}{\beta}}, \quad (6)$$

where $\tilde{y}_i^{(k)}$ indicates the subbands arising from the displayed image $I$, i.e. $f(I) = \{\tilde{y}^{(k)}; k = 1, \ldots, N_k\}$. A similar summation model has been employed in previous perceptual quality metrics [32, 16].

All parameters of the perceptual transform and metric were optimized to best explain human perceptual ratings of distorted images in a public database [24]. Specifically, we chose parameters to maximize the correlation between the mean opinion scores from the human observers and the distance given by out metric. We fixed the parameters prior to using the perceptual model for the rendering results presented below. The front-end nonlinearity used an exponent $\gamma = \frac{1}{2.6}$. Unlike in [15], we set the normalization parameters to be identical for all bandpass channels (assuming scale-invariance), but allowed a different set for the lowpass channel. For bandpass channels, the additive constant was $\sigma = 0.17$, and the local weighting functions $P$ were filters with $5 \times 5$ support, with values

$$P = \begin{bmatrix} 0.04 & 0.04 & 0.05 & 0.04 & 0.04 \\ 0.04 & 0.03 & 0.04 & 0.03 & 0.04 \\ 0.05 & 0.04 & 0.05 & 0.04 & 0.05 \\ 0.04 & 0.03 & 0.04 & 0.03 & 0.04 \\ 0.04 & 0.04 & 0.05 & 0.04 & 0.04 \end{bmatrix}.$$

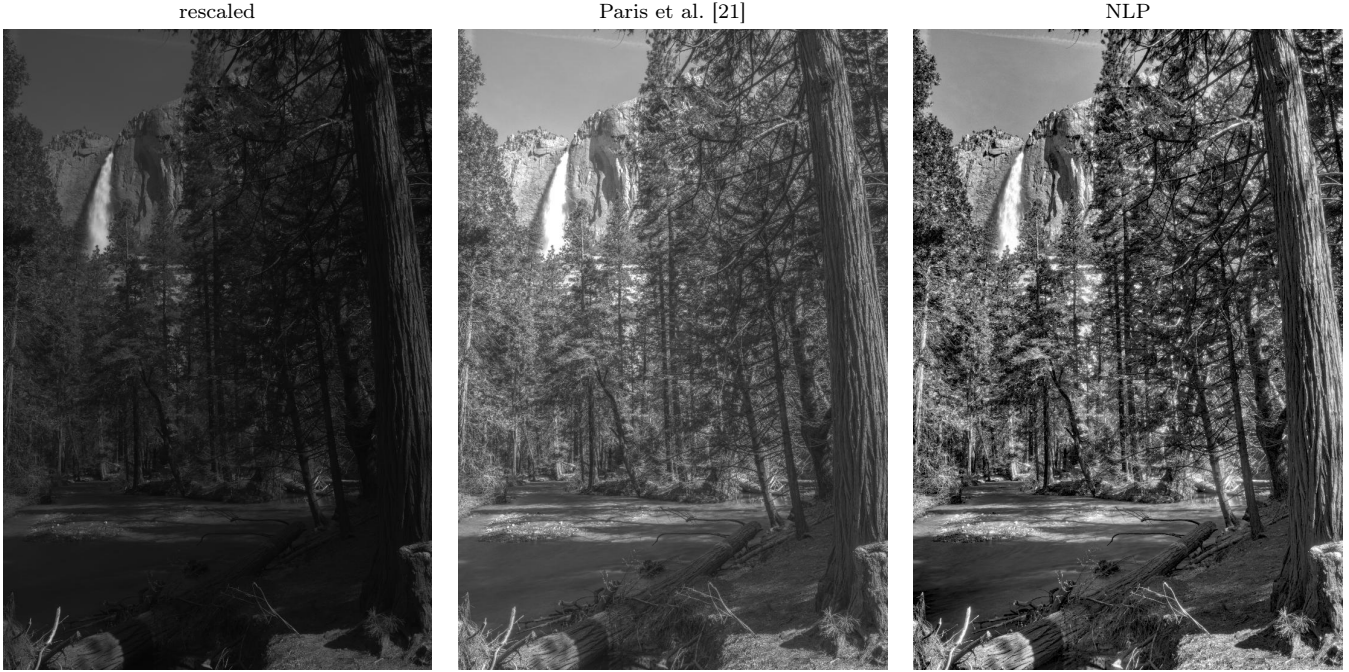

Figure 4: Rendering of a calibrated HDR image on a display with a limited luminance range. The scene luminances for this image spanned the range from $S_{\min} = 0.78\,cd/m^2$ to $S_{\max} = 16200\,cd/m^2$, whereas the display luminances are assumed to lie between $5\,cd/m^2$ and $300\,cd/m^2$. Left: linear rescaling of luminance values into the display range. Center: result obtained from a state-of-the-art tone mapping algorithm [21]. Right: image optimized to minimize perceptual distance while observing the display constraints, using the proposed method.

The parameters for the lowpass channel were $\boldsymbol{P} = 1$ and $\sigma = 4.86$. Optimized exponents for the metric were $\alpha = 2.0$ and $\beta = 0.6$. Appendix B shows that the performance of this extended version of the NLP metric surpasses that of state-of-the-art image quality metrics.

# 3 Varying image acquisition conditions

We performed a set of experiments which demonstrate the flexibility of our optimization framework over different image acquisition conditions. We begin with calibrated images, for which we know the the exact luminance values ($cd/m^2$) in the original scene. We then move on to uncalibrated images, for which we need to make an assumption about the luminance values in the original scene. Finally, we close this section by demonstrating that we can manipulate our assumptions about the original scene luminances to solve other rendering problems, such as haze removal and artificial detail enhancement.

Each example requires us to minimize the perceptual distance with respect to the rendered image $\boldsymbol{I}$, subject to the display constraints. For this, we use the Adaptive Moment Estimation (Adam) algorithm [13]. The derivative of the perceptual distance with respect to $\boldsymbol{I}$ is described in appendix A. The computation time scales linearly with the size of the image. When optimized on a Tesla K40 GPU card, optimization took approximately 1 second for $10,000$ pixels (i.e. an image of $1000 \times 1000$ requires less than 2 minutes). Note that all image results presented here are inteded for viewing on a display with luminance range from 5 to $300\,cd/m^2$, and a gamma exponent of 2.2.

## 3.1 Rendering calibrated HDR luminances

We begin with an image $\boldsymbol{S}$ obtained from an HDR imaging device such that we know the true luminance values of all pixels. The image used here has been extracted from the database of Mark Fairchild [6], and its luminance range is $S_{\min} = 0.78$ to $S_{\max} = 16200\,cd/m^2$. Suppose further that we wish to display this image using a device with a luminance range of $I_{\min} = 5$ to $I_{\max} = 300\,cd/m^2$ (typical for many computer monitors), and that this range is far less than that of the image. We solve for the perceptually optimal rendered image:

$$\hat{\boldsymbol{I}} = \arg\min_{\boldsymbol{I}} D(\boldsymbol{S}, \boldsymbol{I}), \quad \text{s.t. } \forall i : I_{\min} \leq I_i \leq I_{\max}. \quad (7)$$

Results are shown in Fig. 4. We compare the original image intensities, linearly rescaled to fit within the luminance range $[I_{\min}, I_{\max}]$, our perceptually optimized image $\hat{\boldsymbol{I}}$, and an image tone-mapped using a recent state-of-the-art method by Paris et. al. [21]. For the latter, we have used the default parameters recommended by the authors for tone mapping of HDR images: $\alpha = 1$, $\beta = 0$, and $\sigma = \log 2.5$.

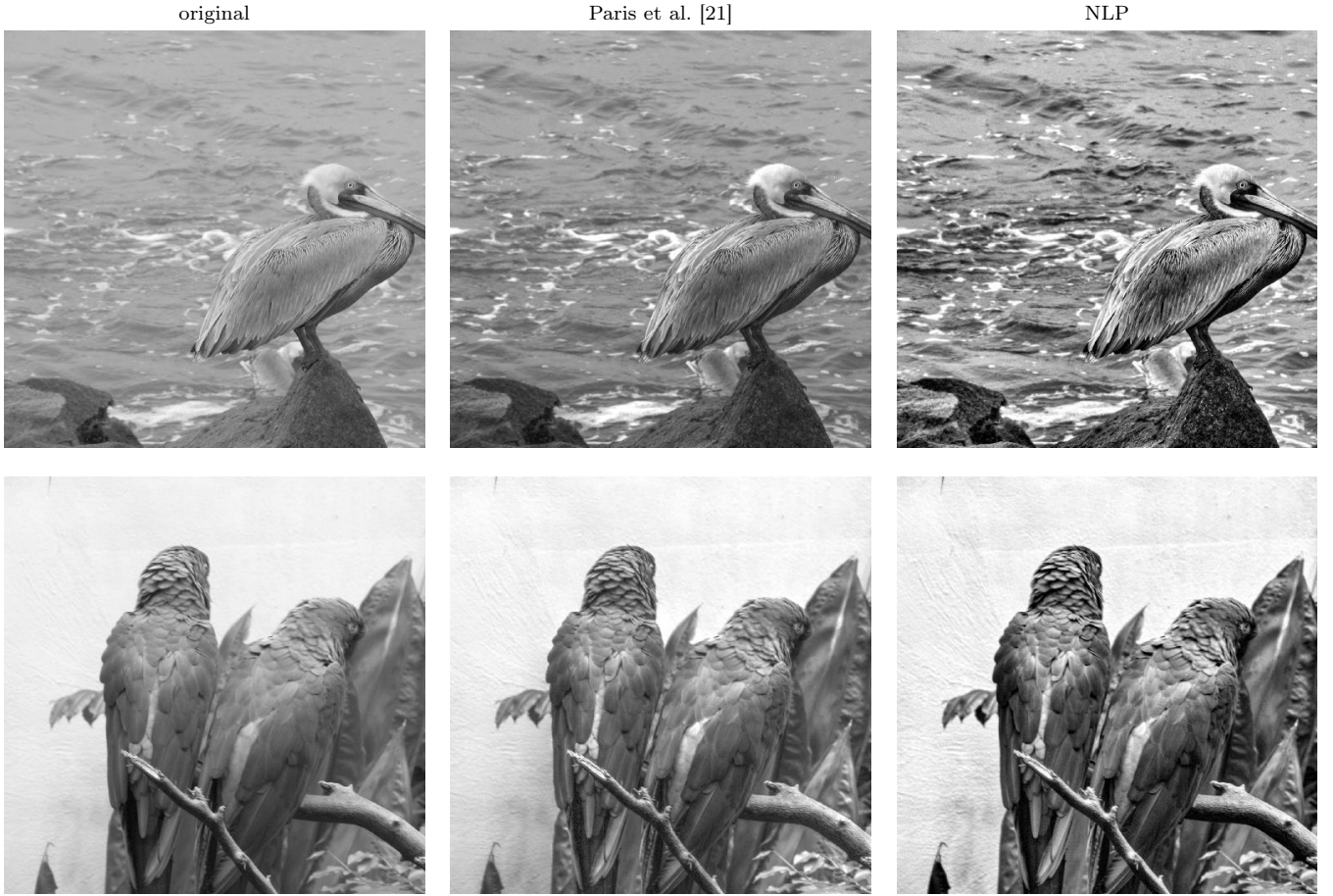

Figure 5: Rendering of calibrated LDR images to a display with a limited luminance range.

Linearly rescaling the values creates a rendered image where most of the the details cannot be seen or differentiated. The algorithm by Paris et. al. [21] mitigates this problem, rendering an image that reveals detail in both dark and bright regions. Nevertheless, the solution appears less detailed and lower in contrast than the image computed using our method. This is mostly because the Paris algorithm does not take into account the display luminance range. Although the algorithm (and most tone-mapping algorithms) has additional parameters that can be adjusted, it is not obvious to a naive user how to select these parameters based on the display properties. In contrast, our solution is fully automatic (assuming the luminance values of the source image and the range of the display are known), albeit at the expense of significantly more computation.

## 3.2 Rendering LDR images with an image acquisition model

Our method can also be used to improve the appearance of images acquired with a conventional low dynamic range (LDR) digital camera that has been calibrated to allow recovery of luminance values (in $cd/m^2$) from recorded pixel values, $\boldsymbol{R}$. For most modern digital cameras, the acquisition luminance range is still generally much larger than the display range, and in any case is unlikely to match. Thus, we need to solve the following optimization problem analogous to the previous section:

$$\hat{\boldsymbol{I}} = \arg\min_{\boldsymbol{I}} D(\boldsymbol{S}, \boldsymbol{I}), \quad \text{s.t.} \quad \forall i : I_{\min} \leq I_i \leq I_{\max} \quad (8)$$

$$\text{where } \boldsymbol{S} = g(\boldsymbol{R}),$$

where $g$ is the mapping from recorded pixel values to estimated scene luminances.

Results for two example grayscale images from the McGill database [20] are shown in Fig. 5. For each image, we again compare the original image intensities, linearly rescaled to fit within the luminance range $[I_{\min}, I_{\max}]$, to our perceptually optimized image $\hat{\boldsymbol{I}}$, and a tone-mapped image computed using the Paris et. al method. [21]. For the latter, we have again used the parameters recommended by the authors for tone mapping of HDR images: $\alpha = 1$, $\beta = 0$, and $\sigma = \log 2.5$. Our method again offers an advantage, producing higher contrast and more detailed results. The improvement here is perhaps even more noticeable than in the HDR case, for which the Paris et. al. algorithm was developed.

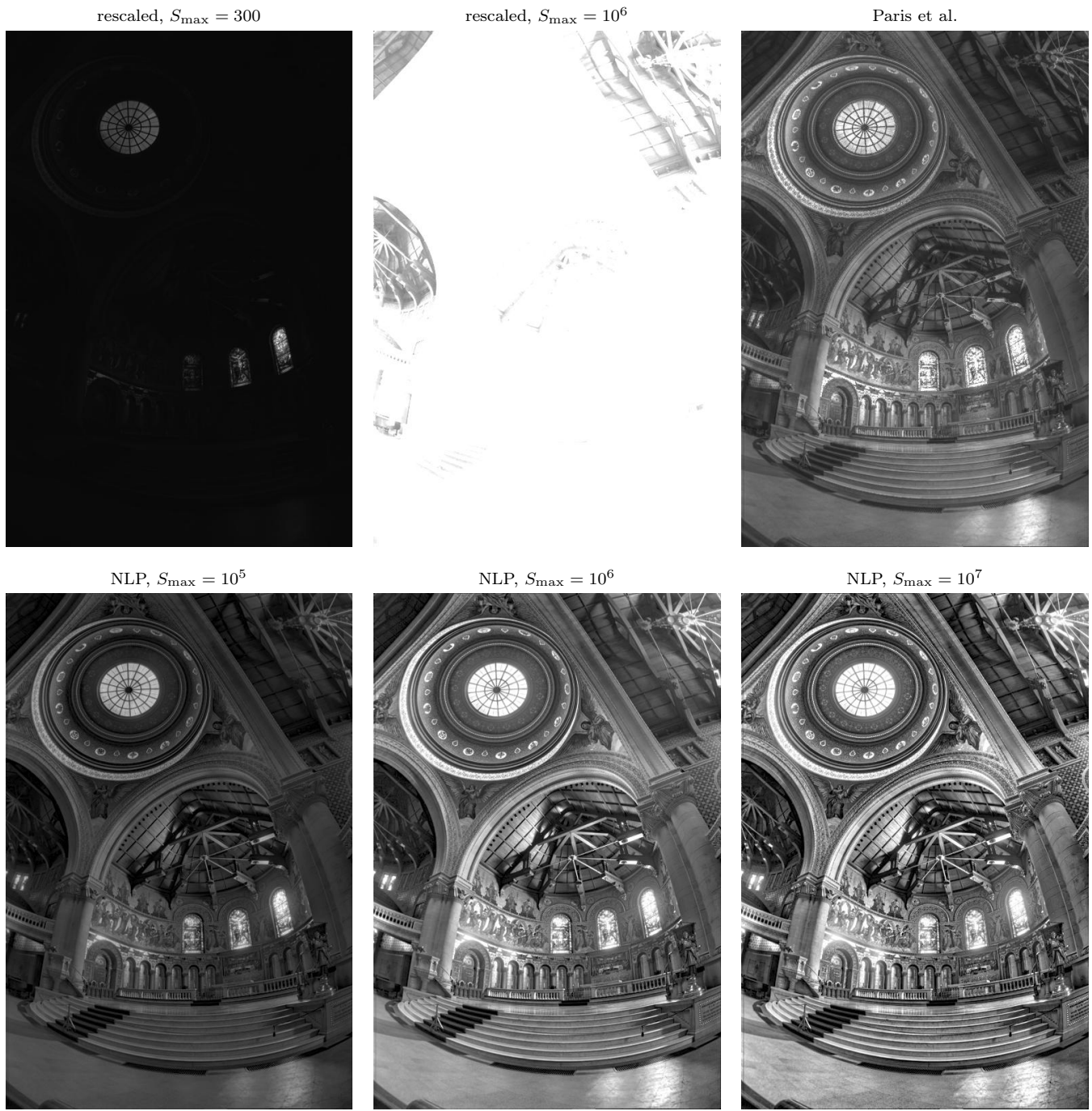

Figure 6: Rendering of an uncalibrated HDR image on a display with a limited luminance range. Linear mapping of luminances leads to loss of detail (top left: rescaling of luminances to the display range corresponds to assuming $S_{\max} = 300\,cd/m^2$; top center: attempting to reproduce luminances one-to-one with realistic assumptions about scene luminance leads to saturation). Top right: Paris et al. [21] result. Bottom: images computed using proposed method, with different assumed maximum luminance values.

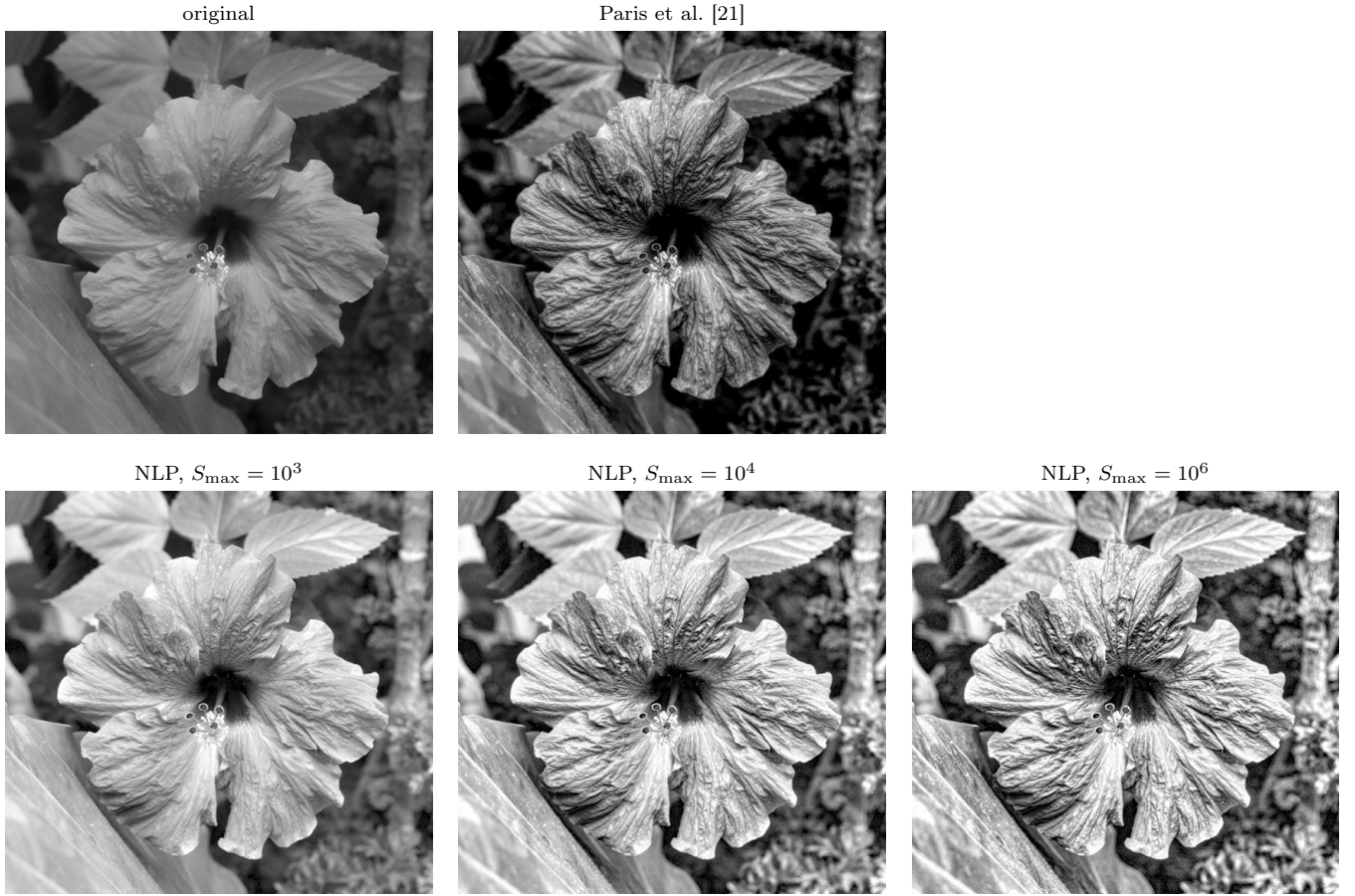

Figure 7: Example of artificial detail enhancement by simulating more light in the original scene. Top left: original image. Top center: image processed using [21]. Bottom: images optimized using NLP and different assumed values of maximum luminance, $S_{\max}$ (we used $S_{\min} = 5$ in all three cases.

## 3.3 Rendering uncalibrated HDR images

Unlike the situation in section 3.1, the typical scenario for images acquired from HDR cameras is that they are uncalibrated. That means that we have access to measurements $\boldsymbol{L}$ that are linearly related to actual luminances, but we do not have access to the scaling parameters (for instance, they might be normalized values, lying between 0 and 1). To apply our method, the measurements need to be linearly rescaled to luminance values, which amounts to knowing, estimating, or guessing the minimum and the maximum luminance in the original scene ($S_{\min}$ and $S_{\max}$, respectively). One can often use an educated guess for those values given the content of the image – for instance, the luminance of a filament of a clear incandescent lamp is roughly $10^6 \, cd/m^2$. As in the previous experiments, we solve the resulting optimization problem:

$$\hat{\boldsymbol{I}} = \arg\min_{\boldsymbol{I}} D(\boldsymbol{S}, \boldsymbol{I}), \quad \text{s.t.} \;\; \forall i : I_{\min} \leq I_i \leq I_{\max} \quad (9)$$

$$\text{where } \boldsymbol{S} = (S_{\max} - S_{\min}) \cdot \boldsymbol{L} + S_{\min}.$$

Figure 6 shows the results for the classical HDR image "Memorial" for different chosen values of $S_{\max}$ (and a fixed value of $S_{\min} = 0.01$). The proposed method converges on an image which exhibits enhanced contrast in all the regions, preserving the details, but also preserving the relative contrast and luminance between regions. This is particularly evident in high luminance regions (for instance the bright window behind the altar, or the round window in the top of the dome), where both the perceived details and luminance intensity is effectively portrayed.

As we increase the maximum luminance chosen for the original scene (while fixing the display restrictions), our algorithm further amplifies the contrast of details in the image. This makes sense from a perceptual point of view. If the original scene was brighter, an observer would be able to perceive more details within the scene. Therefore the method has to artificially enhance these details to mimic the appearance of the original scene. In the next two sections we take advantage of this behavior.

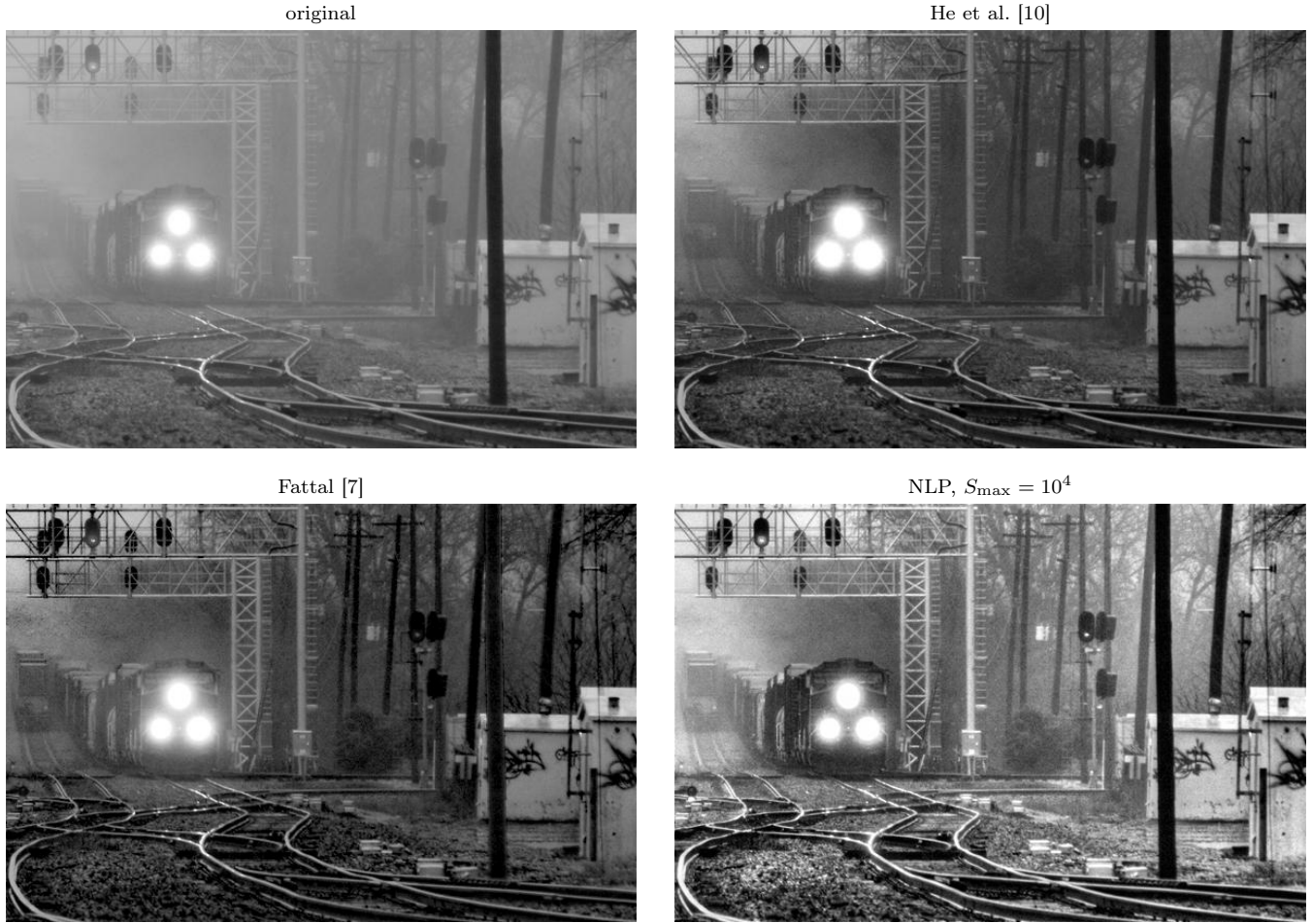

Figure 8: Example of haze removal. Top left: original image. Top right: image processed using He et al. algorithm [10]. Bottom left: image processed using Fattal algorithm [7]. Bottom right: image processed using the proposed method using $S_{\min} = 5$ and $S_{\max} = 10^4$.

### 3.4 Artificial detail enhancement and haze removal

We showed in the preceding sections that using the knowledge we have about the image acquisition process helps greatly in automatically recovering images that are optimized to look like the original scene, given the display constraints. In some cases, however, detail visibility in the scene might be unsatisfactory. Intuitively, photographers know that the amount of detail visible in a scene depends on the amount of available light. If the image has already been acquired, it is of course not possible to alter the light sources. However, since the scene luminances scale linearly with the intensity of the light sources, our method allows us to simulate more light post hoc, by linearly rescaling the scene luminances $\boldsymbol{S}$.

Figure 7 shows the results of modifying our choice of $S_{\max}$ (choosing as in the previous experiment $S_{\min} = 0.01$). Note that with increasing values of $S_{\max}$, details become more visible. For results from the Paris et al. algorithm (shown here as well), we have again employed the parameters proposed in their paper for the detail enhancement problem: $\alpha = 0.25$, $\beta = 1$, and $\sigma = 0.3$.

Surprisingly, this same method of detail enhancement can also be used for the problem of haze removal. In a hazy scene, the local contrast has effectively been reduced (roughly speaking, but adding a constant level of scattered light) which makes detail more difficult to discern. In this experiment, we choose also $S_{\min} = 0.01$ (we find that results are fairly robust to the selection of this parameter) and $S_{\max} = 10^4$. Figure 8 compares the performance of our method with both a classical ([10]) and a state-of-the-art method ([7]). Our algorithm converges on an image that greatly enhances the details of the original hazy image, boosting the contrast and reducing the perception of haze within the image. Although the other two methods are specifically designed for this particular problem, our method obtains a similar result without modification. In this case we used the parameters $S_{\min} = 5$ and $S_{\max} = 10^4$.

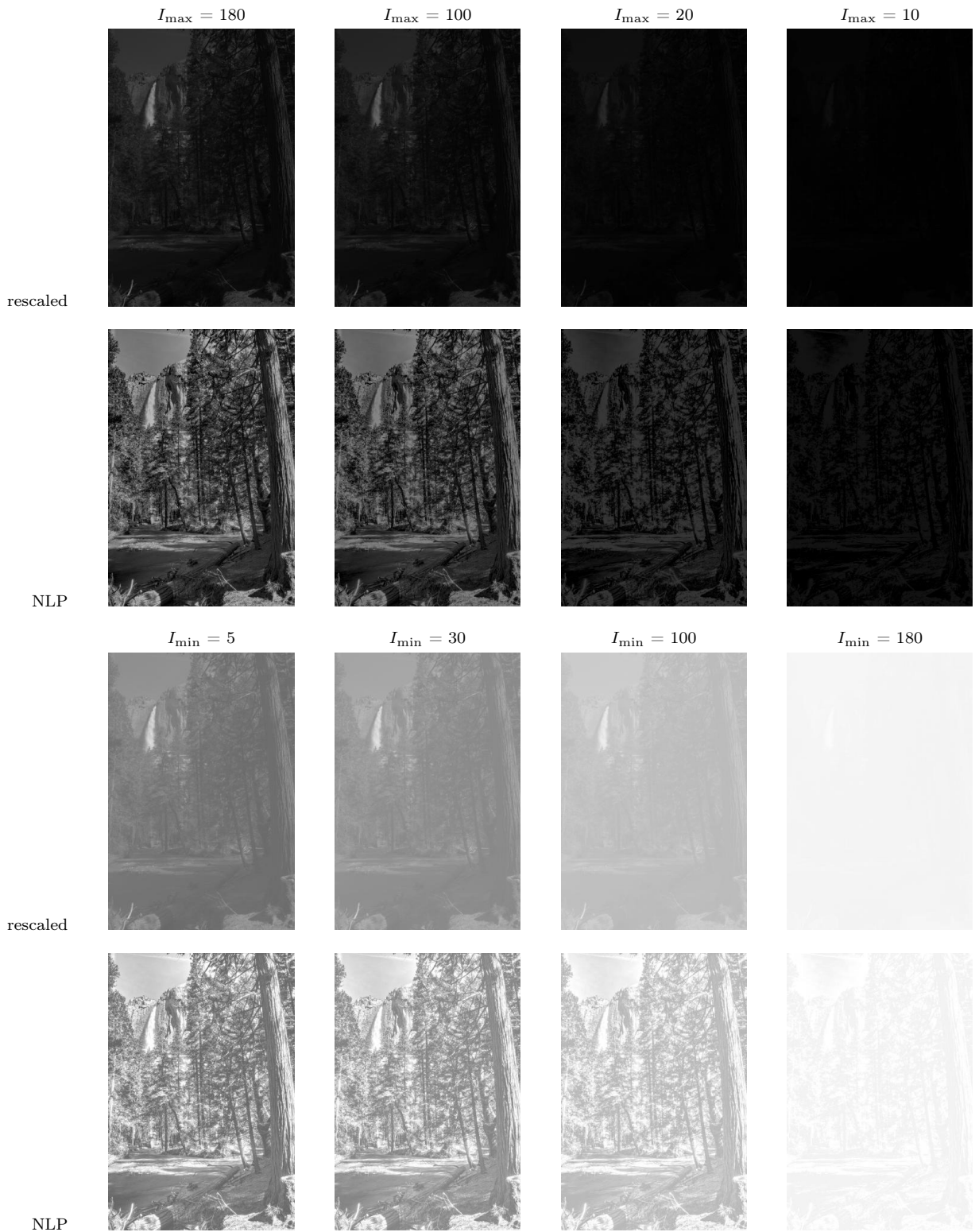

Figure 9: Effect of different maximum and minimum display luminance constraints. Top two rows: Lowered maximum luminance, linearly rescaling the pixel values versus proposed optimization method. Bottom two rows: analogous, but raising minimum luminance.

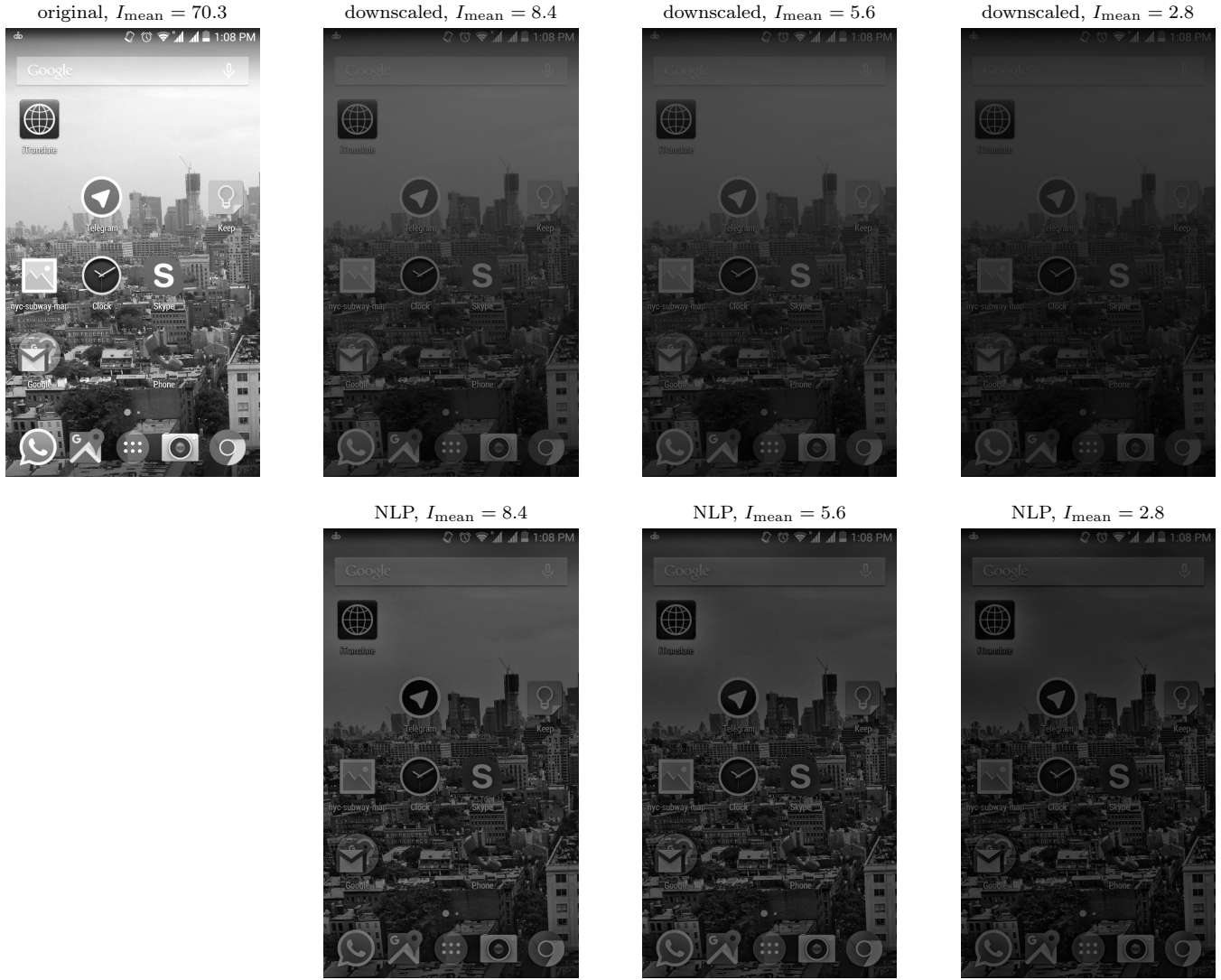

Figure 10: Rendering with an energy consumption constraint. Top left: image at full luminance (smartphone screenshot). Top row: linear rescaling to target mean luminance. Bottom row: images optimized for perceptual distortion with target mean luminance constraint. Note that images in the same column consume the same amount of energy, while the amount of perceptible details is different.

## 4 Varying display constraints

While examining the effects of various image acquisition scenarios in the previous section, we assumed only that the display luminance is bounded. The upper bound is a natural constraint for any real display. The lower bound is also relevant for a wide range of practical display devices, and arises from reflected ambient light and scatter within the display. In this section, we examine the effect of each of these constraints independently, along with a few more complex constraints.

Figure 9 shows the results for different maximum ($I_{\max}$) and minimum ($I_{\min}$) luminance bounds. Our method enhances local contrast, whereas a rescaling can only manipulate contrast globally. For a wide range of display characteristics, optimizing the image to minimize the NLP distance reduces distortion in the rendered images, and increases the visibility of perceptually relevant features.

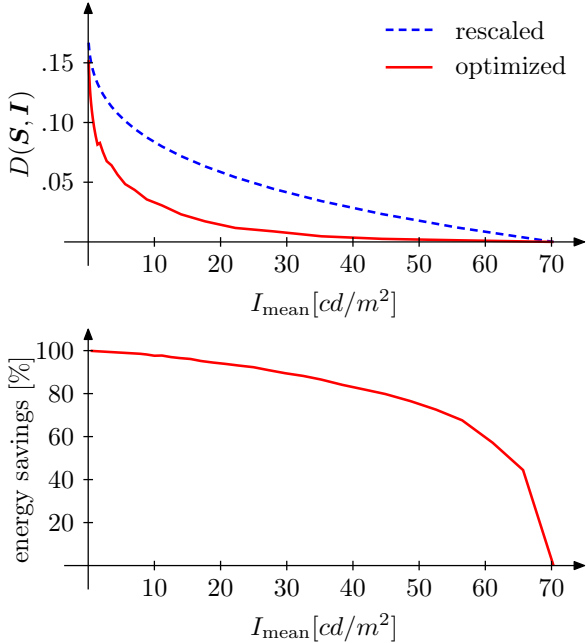

Figure 11: Tradeoff between energy consumption and image quality. Top: perceptual distortion as a function of mean luminance (assumed proportional to energy consumption), for linearly rescaled images vs. those optimized for perceptual distortion with a mean luminance constraint. Bottom: fraction of energy saved with optimized solution compared to linear rescaling, as a function of mean luminance.

## 4.1 Rendering with limited energy consumption

The proposed framework allows us to seamlessly introduce arbitrary display constraints. For example, we can optimize the tradeoff between image quality and energy consumption. We assume that energy consumption is proportional to luminance, as for instance in organic light-emitting diode (OLED) displays used in cell phones. Thus, we keep the luminance within a lower and upper bound, while also keeping the mean luminance constant:

$$\hat{\boldsymbol{I}} = \arg\min_{\boldsymbol{I}} D(\boldsymbol{S}, \boldsymbol{I}), \quad \text{s.t.} \ \forall i : I_{\min} \leq I_i \leq I_{\max} \quad (10)$$

$$\text{and} \ \frac{1}{N_i}\sum_i I_i = I_{\text{mean}}$$

Figure 10 shows images optimized for different mean luminance compared to images linearly rescaled to achieve the same target mean luminance. It is clear that our optimized images retain more detail from the original scene. Figure 11 shows two plots where the visual appearance and the energy consumption are compared for both methods. Optimizing the images yields a clear benefit in terms of the tradeoff between consumed energy versus perceptual distortion.

## 4.2 Rendering with a discrete set of gray levels (dithering)

Most displays have a limited number of available gray levels. In the extreme case this can be as few as two (e.g., black-and-white printers, e-ink devices, etc). Here, we illustrate that the proposed method is flexible enough to produce good results even under such extreme constraints. The optimization problem is the same as before, but here, we restrict the pixel values to be taken from a discrete set:

$$\hat{\boldsymbol{I}} = \arg\min_{\boldsymbol{I}} D(\boldsymbol{S}, \boldsymbol{I}), \quad \text{s.t.} \ \forall i : I_i \in \{I_{\min}, \ldots, I_{\max}\}. \quad (11)$$

The discrete nature of the optimization problem prevents us from using a gradient-based method for optimization. Instead, we use a greedy error-diffusion algorithm, analogous to the classic Floyd–Steinberg method. We first initialize the image to the continuous solution obtained for a continuous range of luminances, as in previous experiments. Then, we iteratively select the discrete value for each pixel of the image in raster-scan order, each time picking the discrete value that minimizes the NLP distance of the intermediate result to the original scene.

Figure 12 shows the results for images rendered using two and four gray levels. In low contrast regions, our method is seen to preserve significantly more detail than the Floyd–Steinberg method. In addition, the Floyd–Steinberg algorithm tends to return artificially imposed patterns in long flat regions, which can be seen in the dark regions of the penguin's wings. Our method, however, does not generate these artificial patterns.

## 5 Contribution of perceptual metric components

In order to provide intuition regarding the effect of each of the primary components of the NLP, we optimized images for rendering while removing one of three components of the transform: the initial pointwise nonlinearity (set $\gamma = 1$), the multi-scale decomposition (set $N_k = 1$), and divisive normalization (set $P = 0$ and $\sigma = 1$). Figure 13 shows results for each manipulation. Note that we did not refit each of the partial transforms to predict human perceptual judgments; therefore, these results should be seen as a way to understand the importance of each computation, and not as a quantitative comparison of image quality assessment performance (see details in appendix B).

Each of the three images differs noticeably from the one optimized with the full transform. Without the initial pointwise nonlinearity, the algorithm produces images in which low to medium luminance patches of an image are misrepresented. The high luminance areas are detailed but some parts with medium or low luminance are almost flat. Without the multi-scale decomposition, the algorithm produces images in which extremely high and extremely low frequencies are well preserved, but intermediate frequencies are underrepresented, and some cases nearly disappear. Without the normalization, the algorithm converges to images that saturate at the luminance boundary constraints of the display. Normalization preserves the relative luminance changes between coefficients while allowing the absolute luminance to be modified. This allows the rendered image pixel intensities to be proportional to the relative energy in each local region. Moreover, this ensures that regions with similar content scale in a similar way.

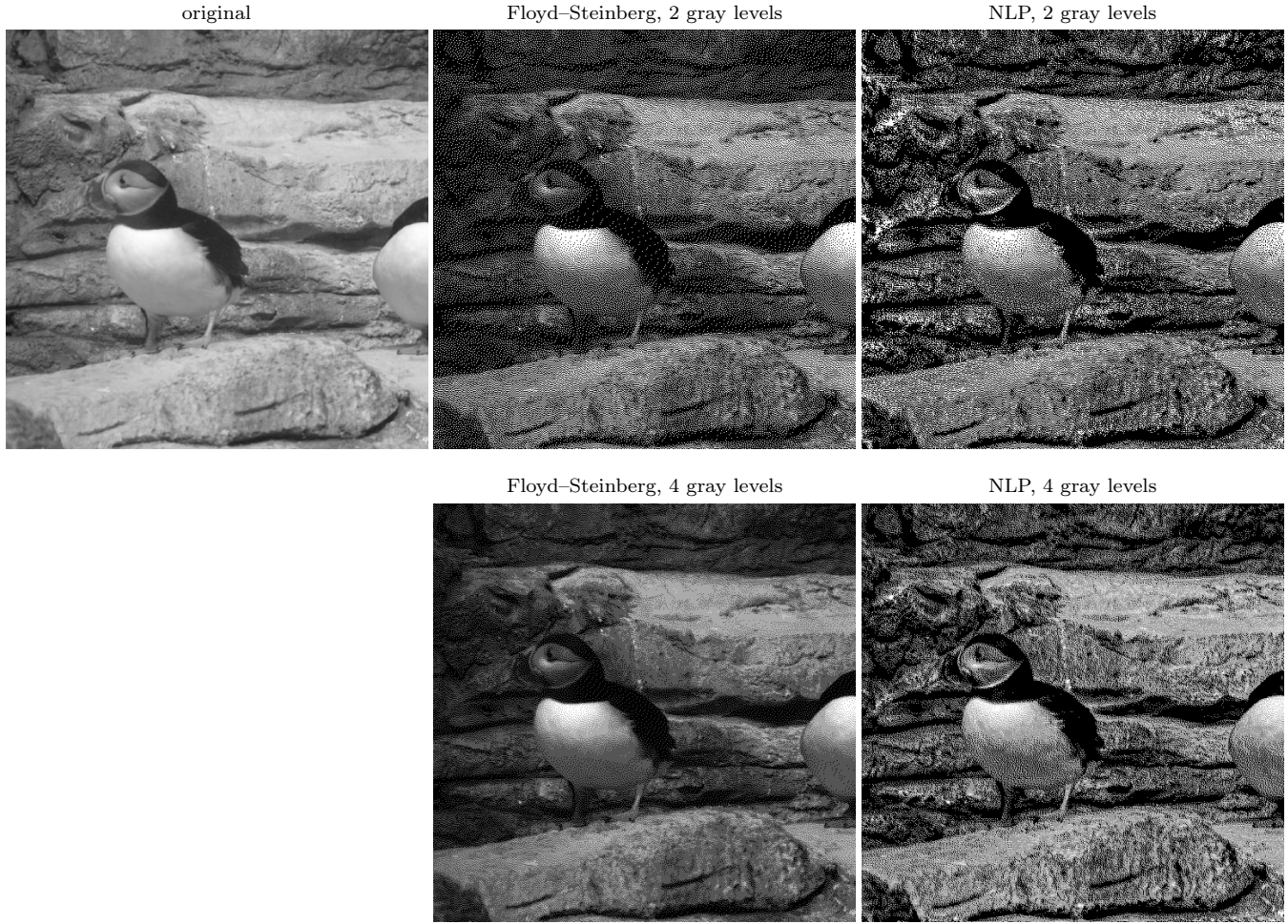

Figure 12: Rendering with a discrete set of gray levels. Top left: original image. Center column: Floyd–Steinberg method [9]. Right column: optimized using NLP.

## 6 Discussion

We've described a method of directly optimizing rendered images, so as to minimize their perceptual difference from the original scene from which they were derived. Perceptual optimization of tone mapping algorithms is not a new concept. For example, Tumblin and Rushmeier's seminal paper on tone mapping states: "Accurate display methods should compensate for all light dependent changes in the way we see" [29]. The authors propose optimizing a tone mapping operator (i.e. the transformation from HDR to LDR) to best match the appearance between the displayed image and the scene. Several tone mapping papers have followed this framework (see for instance [8, 22, 28, 23, 18]), each using a different perceptual metric to determine the similarity between the rendered image and the scene. These methods depend critically on the particular parametric function they use as a tone mapping operator, which restricts the space of possible solutions. Thus, a given functional form may not allow the optimal solution to be found, or may only work satisfactorily for a particular type of rendering constraint.

Nowadays, tone mapping methods often do not make explicit use of perceptual metrics (see [5] for a nice review), but rather provide the user with a small set of free parameters to hand-control the mapping from the scene to the displayed image. These methods are conceptually simpler than ours, and some of them can produce high quality results in controlled situations (see for instance [21]). Nevertheless, their parameter are often difficult to interpret (and thus, to set), and the restriction to particular functional forms may limit their applicability to different rendering problems.

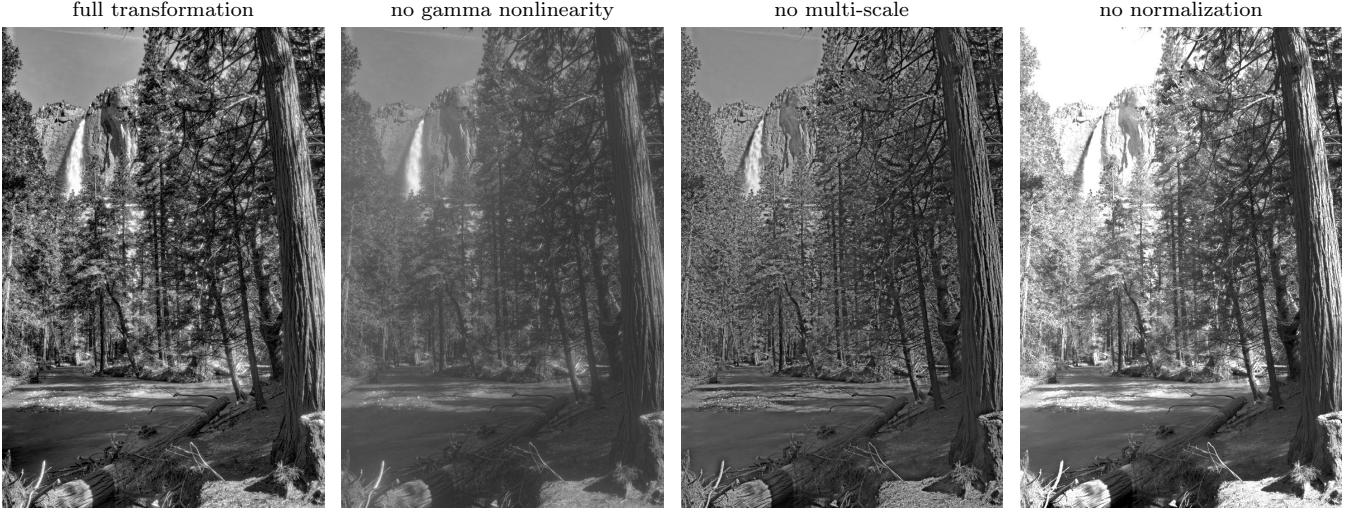

Figure 13: Rendering of an HDR image when removing different parts of the NLP transformation (see text).

In contrast, by directly optimizing the rendered image itself, our method is free to take into account different display constraints, without requiring manual selection of an appropriate form for each situation. The downside of this approach is computational cost: optimization over the high-dimensional space of rendered images is expensive, and although both hardware and software continue to improve, it will presumably always be significantly more expensive than optimizing a small set of parameters for a fixed transformation. But even if the computational costs prevent the use of this method in a given application, the results can still serve as a benchmark for what is possible, thus facilitating the development of alternative methods.

We've employed a perceptual metric based on an abstraction of the physiological transformations implemented in the early stages of the visual system. The metric is an extension of the NLP distance presented in [15]. We have show that this metric is consistent with human perception, exhibiting correlation to human quality ratings that is similar to or better than models specifically designed to assess perceptual quality (see appendix B). It is continuous and differentiable, with well-behaved gradients, making it easy to incorporate into optimization procedures. As a case in point, it has also been employed to optimize an image compression algorithm [1].

Although our framework may be applied to any display problem, it depends heavily on both the HVS model employed, and the method used to solve the constrained optimization problem. There is room for improvement in both of these elements. Optimization has undergone dramatic changes in the past decade, and methods for handling nonconvex and even discrete problems have become more reliable and efficient. In particular, it should be possible to improve the halftoning solution, for which we used a simple greedy method analogous to Floyd-Steinberg [9].

Our use of a simple physiologically-inspired model for assessing perceptual disortion offers opportunities for improvement (note that most image quality models are less physiologically based [30, 31, 19]). For example, the NLP can likely be improved by including relationships between channels, which could help to control artifacts such as halos that sometimes appear around edges with large contrast. In addition, the NLP model should be extended to operate on color images. This can be done following the same procedure that we used for the achromatic model: taking inspiration from human physiology to define the functional form, and fitting the parameters using human psychophysical data. It should also be beneficial to extend the model to include another stage of processing corresponding to primary visual cortex, and containing oriented, multi-scale, derivative filters.

# A  Derivative of the distance with respect to the rendered image

Here we provide, for the interested reader, the derivative of the perceptual distance $D(\boldsymbol{S}, \boldsymbol{I})$ with respect to the rendered image $\boldsymbol{I}$. The distance is given by:

$$D(\boldsymbol{S}, \boldsymbol{I}) = \left[ \frac{1}{N_k} \sum_{k=1}^{N_k} \left( \frac{1}{N_c^{(k)}} \sum_{i=1}^{N_c} \left(d_i^{(k)}\right)^\alpha \right)^{\frac{\beta}{\alpha}} \right]^{\frac{1}{\beta}}$$

Here, we summarized the distance between the transformed images, $y = f(\boldsymbol{S})$ and $\tilde{y} = f(\boldsymbol{I})$, as:

$$d_i^{(k)} = |y_i^{(k)} - \tilde{y}_i^{(k)}|$$

The general equation is:

$$\frac{\partial D(\boldsymbol{S}, \boldsymbol{I})}{\partial \boldsymbol{I}_j} = \frac{1}{\beta} D(\boldsymbol{S}, \boldsymbol{I})^{1-\beta} \frac{\partial}{\partial \boldsymbol{I}_j} \left[ \frac{1}{N_k} \sum_{k=1}^{N_k} \left( \frac{1}{N_c^{(k)}} \sum_{i=1}^{N_c} d_i^{(k)\alpha} \right)^{\frac{\beta}{\alpha}} \right]$$

From here we only apply the chain rule to expand the full equation. The second term in the equation above is:

$$\frac{\partial}{\partial \boldsymbol{I}_j} \left[ \frac{1}{N_k} \sum_{k=1}^{N_k} \left( \frac{1}{N_c^{(k)}} \sum_{i=1}^{N_c} d_i^{(k)\alpha} \right)^{\frac{\beta}{\alpha}} \right] =$$

$$\frac{\beta}{\alpha N_k} \sum_{k=1}^{N_k} \left( \frac{1}{N_c^{(k)}} \sum_{i=1}^{N_c} d_i^{(k)\alpha} \right)^{\frac{\beta}{\alpha}-1} \frac{\partial}{\partial \boldsymbol{I}_j} \left[ \frac{1}{N_c^{(k)}} \sum_{i=1}^{N_c} d_i^{(k)\alpha} \right]$$

And we can put the second term in the equation above in function of the derivative of the difference:

$$\frac{\partial}{\partial \boldsymbol{I}_j} \left[ \frac{1}{N_c^{(k)}} \sum_{i=1}^{N_c} d_i^{(k)\alpha} \right] = \frac{\alpha}{N_c^{(k)}} \sum_{i=1}^{N_c} d_i^{(k)(\alpha-1)} \frac{\partial d_i^{(k)}}{\partial \boldsymbol{I}_j}$$

By putting all terms together we have the derivative $\frac{\partial D(\boldsymbol{S}, \boldsymbol{I})}{\partial \boldsymbol{I}_j}$ as a function of the derivative of the difference:

$$D(\boldsymbol{S}, \boldsymbol{I})^{1-\beta} \frac{1}{N_k} \sum_{k=1}^{N_k} \frac{1}{N_c^{(k)\frac{\beta}{\alpha}}} \left( \sum_{i=1}^{N_c} d_i^{(k)\alpha} \right)^{\frac{\beta}{\alpha}-1} \sum_{i=1}^{N_c} d_i^{(k)(\alpha-1)} \frac{\partial d_i^{(k)}}{\partial \boldsymbol{I}_j}$$

And now expanding the derivative of the difference:

$$\frac{\partial d_i^{(k)}}{\partial \boldsymbol{I}_j} = \text{sgn}(y_i^{(k)} - \tilde{y}_i^{(k)}) \frac{\partial y_i^{(k)}}{\partial \boldsymbol{I}_j}$$

$$\frac{\partial y_i^{(k)}}{\partial \boldsymbol{I}_j} = \left( \frac{\partial y_i^{(k)}}{\partial \boldsymbol{z}^{(k)}} \right) \left( \frac{\partial \boldsymbol{z}^{(k)}}{\partial x_j} \right) \left( \frac{\partial x_j}{\partial \boldsymbol{I}_j} \right)$$

The derivative for the first term is:

$$\frac{\partial y_i^{(k)}}{\partial z_i^{(k)}} = \frac{\sigma + \boldsymbol{P}_i |\boldsymbol{z}| - P_{ii} \text{sgn}(z_i) z_i}{(\sigma + \boldsymbol{P}_i |\boldsymbol{z}|)^2}$$

$$\frac{\partial y_i^{(k)}}{\partial z_l^{(k)}} = \frac{-P_{il} \text{sgn}(z_l) z_i}{(\sigma + \boldsymbol{P}_i |\boldsymbol{z}|)^2}, l \neq i$$

Where $sgn(z_i)$ is the sign of $z_i$. The second term is:

$$\frac{\partial \boldsymbol{z}^{(k)}}{\partial x_j} = \boldsymbol{Q}_{(:,j)}^{(k)}$$

Where $\boldsymbol{Q}$ is the matrix of the linear transformation performed by the Laplacian Pyramid, $\boldsymbol{z} = \boldsymbol{Q}\boldsymbol{x}$. The third term is:

$$\frac{\partial x_j^{(k)}}{\partial \boldsymbol{I}_j} = \frac{1}{\gamma} \boldsymbol{I}_j^{(\frac{1}{\gamma}-1)}$$

# B  IQA performance of Normalized Laplacian Pyramid

Perceptual image quality assessment (IQA), as a means of comparing results obtained by different methods, has become an important topic in image processing. Although the best method of evaluating IQA is through explicit measurement of human responses, this is a difficult and costly undertaking. An objective measure of perceptual quality alleviates this difficulty. If the measure is differentiable and well-behaved, additional advantage arises from using it to optimize the perceptual performance of algorithms.

The most widely-used method of assessing IQA models is by measuring their correlation with human quality ratings on a diverse set of distorted images [30, 31, 19]. Table 1 presents correlation results against five databases of human mean opinion scores: four were measured using low dynamic range displays, and one was targeted at HDR displays. All results are obtained using the achromatic images in the databases (we have not yet extended the NLP metric to handle color), and we also include results for several widely employed IQA methods. The table shows results for two types of correlation: The Pearson correlation, which measures linear predictability of the human responses, and the Spearman correlation, which is concerned only with the ranking of the responses, and thus more robust to (monotonic) nonlinear distortions. Note that the latter measure is perhaps too flexible, since the nonlinear relationship between the MOS and the predicted value can be different for each database, it is often reported when evaluating IQA methods, and we include it here for completeness.

Our results indicate that the proposed metric behaves well for both low and high dynamic range images. Note that the parameters of our metric were adjusted using the TID 2008 [24] database, the VDP 2.2 metric was trained using HDR images, the TID 2008 [24] and the CSIQ [17] database, and the SSIM and MS-SSIM were trained using LIVE database [27]. The Pearson (linear) correlation of our proposed metric is clearly better for four of five datasets (including the HDR dataset), and the Spearman (nonlinear) correlation is equal to or better than all the other metrics for all the datasets. We conclude that our proposed NLP metric is competitive with the current state-of-the-art in IQA. In addition, the NLP metric is the only one that has shown to be easily differentiable for incorporation into convex optimization procedures. For example, the application of SSIM to optimization procedures is not straightforward and it involves some modifications of the metric [2].

Table 1: Evaluation of IQA methods in different databases. Pearson correlation and Spearman correlation (in parentheses) of distance metrics vs. human perceptual judgments.

|         | TID 2008 [24] |        | TID 2013 [25] |        | LIVE [27] |          | CSIQ [17] |        | EPFL [14] |        |
|---------|---------------|--------|---------------|--------|-----------|----------|-----------|--------|-----------|--------|
| PSNR    | 0.52          | (0.55) | 0.45          | (0.64) | 0.86      | (0.94)   | 0.76      | (0.81) | 0.78      | (0.79) |
| SSIM    | 0.74          | (0.78) | 0.76          | (0.74) | 0.83      | (**0.97**) | 0.79      | (0.87) | 0.75      | (0.95) |
| MS-SSIM | 0.79          | (0.85) | 0.78          | (0.79) | 0.77      | (**0.97**) | 0.77      | (0.91) | 0.79      | (0.94) |
| VDP 2.2 | 0.80          | (0.85) | 0.66          | (0.77) | **0.93**  | (**0.97**) | **0.90**  | (0.92) | 0.90      | (0.95) |
| NLPD    | **0.89**      | (**0.89**) | **0.83**  | (**0.80**) | 0.89  | (**0.97**) | **0.90**  | (**0.93**) | **0.93** | (**0.96**) |

# Acknowledgments

JB and EPS are supported by the Howard Hughes Medical Institute. VL is supported by the APOSTD/2014/095 Generalitat Valenciana grant (Spain) and Analog Devices, Inc. AB is supported by the NEI Visual Neuroscience Training Program, T32 EY007136. We want to thank the comments of many people that have helped during the development of this work, in particular Jesús Malo, Javier Calpe, Pau Seguí, Jorge Pérez, Marcelo Bertalmío, Ted Adelson, Alejandro Párraga, Xim Cerdá, Sylvian Paris, Mark Fairchild, and all the people from LCV.